\documentclass[10pt,twocolumn]{article}

\usepackage[margin=1in]{geometry}
\usepackage{graphicx}
\usepackage{booktabs}
\usepackage{amsmath}
\usepackage{amssymb}
\usepackage[authoryear,sort&compress]{natbib}
\usepackage{hyperref}
\usepackage{xcolor}
\usepackage{microtype}
\usepackage{enumitem}
\usepackage{tabularx}
\usepackage{multirow}
\usepackage{float}
\usepackage[font=small,labelfont=bf]{caption}

\hypersetup{
  colorlinks=true,
  linkcolor=blue!60!black,
  citecolor=blue!60!black,
  urlcolor=blue!60!black
}

\bibpunct{(}{)}{;}{a}{,}{,}

\title{Predictive Associative Memory: Retrieval Beyond Similarity Through Temporal Co-occurrence}

\author{Jason Dury\\Independent Researcher\\jason@eridos.ai}

\date{}

\begin{document}

\maketitle


\begin{abstract}
Current approaches to memory in neural systems rely on similarity-based retrieval: given a query, find the most representationally similar stored state. This assumption --- that useful memories are similar memories --- fails to capture a fundamental property of biological memory: association through temporal co-occurrence. Stairs do not resemble a slip, yet one reliably evokes the other. We propose \emph{Predictive Associative Memory} (PAM), an architecture in which a JEPA-style predictor, trained on temporal co-occurrence within a continuous experience stream, learns to navigate the associative structure of an embedding space. We show that this single mechanism --- predicting which regions of meaning space are reachable from a given state --- provides a unified predictive substrate for associative recall, and a plausible architectural basis for imagination and creative recombination. We introduce an \emph{Inward JEPA} that operates over stored experience (predicting associatively reachable past states) as the complement to the standard \emph{Outward JEPA} that operates over incoming sensory data (predicting future states). We evaluate PAM as an \emph{associative recall} system --- testing faithfulness of recall for experienced associations --- rather than as a retrieval system evaluated on generalisation to unseen associations. On a synthetic benchmark, the predictor's top retrieval is a true temporal associate 97\% of the time (Association Precision@1 = 0.970); it achieves cross-boundary Recall@20 = 0.421 where cosine similarity scores zero; and it separates experienced-together from never-experienced-together states with a discrimination AUC of 0.916 (cosine: 0.789). Even restricted to cross-room pairs where embedding similarity is uninformative, the predictor achieves AUC = 0.849 (cosine: 0.503, chance). Specificity controls confirm the predictor discriminates true temporal associates from similar-but-not-associated distractors (AUC = 0.848 vs cosine 0.732). A temporal shuffle control confirms the signal is genuine temporal co-occurrence structure, not embedding geometry: shuffling temporal order within trajectories collapses cross-boundary recall by 90\%, replicated across training seeds. A held-out query-state evaluation confirms anchor-specific recall: the predictor recalls associations from experienced viewpoints but not from novel ones, consistent with the episodic character of the memory system. All primary results are stable across training seeds (SD $< 0.006$) and query selections (SD $\leq 0.012$). We propose that the interaction between similarity (Outward) and association (Inward) would produce episodic specificity --- ``that drill that stripped the screw on Tuesday'' --- that neither channel provides alone; the present work validates the Inward channel in isolation.
\end{abstract}


\section{Introduction}

Consider the experience of walking down a staircase that is slightly damp. Before any conscious analysis, you feel a flicker of unease --- a memory of slipping, perhaps years ago on different stairs entirely. The retrieved memory bears no representational similarity to the current scene. The lighting, the staircase, your clothing, the decade --- all different. What links them is not what they look like, but that they were temporally adjacent to the same kind of event: the moment before a fall.

This is a pervasive mode of episodic memory recall. The hippocampal indexing theory \citep{teyler1986hippocampal} holds that the hippocampus binds cortical representations that were active within the same temporal window, creating associative links between experiences that may share no featural similarity. Complementary learning systems theory \citep{mcclelland1995cls} formalises the division: the neocortex learns slow statistical regularities (similarity structure), while the hippocampus rapidly encodes the specific conjunctions of a single episode (associative structure). The smell of sunscreen retrieves a beach holiday. A chord progression retrieves a person. A particular quality of afternoon light retrieves a childhood kitchen. In every case, the retrieved memory is linked to the current experience not by similarity in any representable feature space, but by temporal co-occurrence --- having been experienced within the same temporal window as something structurally related to the present. This is not incidental to biology. Any agent that experiences reality sequentially will produce temporally ordered memory traces, making temporal co-occurrence a universal associative signal that requires no annotation, no supervision, and no representational overlap between linked experiences.

Despite this, the dominant paradigm for memory in modern neural architectures is similarity-based retrieval. Retrieval-Augmented Generation (RAG) encodes documents and queries into a shared embedding space and retrieves by cosine similarity or approximate nearest neighbours \citep{johnson2019billion,lewis2020rag}. Memory Networks learn to address stored memories, but the addressing mechanism is trained end-to-end on task loss --- the associative structure is a side effect of optimisation, not learned from the structure of experience itself \citep{weston2015memory,sukhbaatar2015e2e}. Modern continuous Hopfield networks provide elegant pattern completion within stored memories but encode no structure \emph{between} memories \citep{krotov2016dense,ramsauer2021hopfield}. Knowledge graphs capture relational structure but require it to be hand-built or extracted by a separate system \citep{bordes2013transe}.

The closest conceptual ancestor to what we propose is spreading activation \citep{collins1975spreading}, the cognitive science model in which activating one node in a semantic network spreads activation to associated nodes. Collins and Loftus had the right idea about traversal of associative structure. However, spreading activation was implemented as traversal of hand-built semantic graphs with manually assigned link weights. It described the phenomenon without providing a mechanism for \emph{learning} the associative structure from experience, or for \emph{implementing} the traversal as a differentiable computation.

We propose to close this gap. Predictive Associative Memory (PAM) is an architecture in which:

\begin{enumerate}[leftmargin=*]
\item A \textbf{JEPA-style encoder} maps continuous multimodal experience into a shared embedding space --- composite states capturing visual, auditory, proprioceptive, and contextual information at each moment.

\item An \textbf{Inward predictor} is trained on temporal co-occurrence: given state $s(t)$, predict the region of embedding space containing states from the temporal neighbourhood of $s(t)$ across the agent's entire history. The training signal is \emph{reachability through learned associative structure}.

\item \textbf{Memory retrieval} is a forward pass through the predictor: the output region specifies which stored states are associatively reachable from the current moment.

\item \textbf{Imagination} uses the same mechanism to reach regions beyond any stored state, grounded in actually-experienced transition patterns.

\item \textbf{Creative recombination} would emerge when the predictor bridges regions linked only by transitive chains: trajectory 1 visits states A$\rightarrow$B, trajectory 2 visits B$\rightarrow$C. Given A, the predictor could reach C --- a connection never directly experienced, but every link grounded in actual temporal co-occurrence. This requires entity persistence across episodes, which our benchmark does not yet provide (Section~5.5).
\end{enumerate}

The Inward predictor is the complement of the standard JEPA predictor (which we term the \emph{Outward} predictor). The Outward predictor, given current sensory state, predicts the next sensory state --- it faces forward into real time. The Inward predictor, given current state, predicts which past states are experientially reachable --- it faces sideways into memory time. Same architecture, same prediction-in-latent-space mechanism, different target space. The claim is that memory retrieval, forward prediction, imagination, and creative recombination are all instances of the same computation: predicting a target region in meaning space given a source state.

This unification has a concrete architectural consequence: a system with both Outward and Inward predictors would achieve \emph{episodic specificity} through the interaction of similarity and association. The Outward encoder learns functional similarity (``these things share properties --- drills and impact drivers''). The Inward predictor learns experiential association (``these things were experienced together --- this drill and this specific screw on Tuesday''). When both converge on the same target, the search space is divided, producing the specificity that characterises vivid episodic memory. The present work validates the Inward channel in isolation; the dual-channel interaction is an architectural prediction.

In this paper, we formalise the architecture, position it against five families of related work, and present results from a synthetic benchmark designed to evaluate the \emph{faithfulness of associative recall} --- how accurately the predictor recalls associations that were actually experienced --- rather than generalisation to unseen associations. We draw a distinction between \emph{retrieval} (finding relevant items from a corpus, where generalisation is the goal) and \emph{recall} (re-activating specific associations formed by specific experience, where faithful memorisation is the goal). The benchmark evaluates: (1)~cross-boundary recall of associations that span representational boundaries, (2)~discrimination between experienced and non-experienced associations, (3)~episodic specificity against similar-but-not-associated distractors, and (4)~validation that the learned signal reflects genuine temporal structure rather than embedding geometry artifacts. We additionally report a negative result on creative bridging that clarifies the boundary conditions for cross-episode recombination.


\section{Related Work}

We organise related work into five families, each capturing a different approach to memory in neural systems. Our contribution sits in the gap between them: learned associative structure from temporal co-occurrence, no explicit graph construction, no task supervision, and a neural predictor that provides the traversal mechanism that spreading activation models describe but never implemented as a learned, differentiable operation.

\subsection{Similarity-Based Retrieval}

The dominant paradigm for memory in deployed systems is retrieval by embedding similarity. Retrieval-Augmented Generation (RAG) encodes documents and queries into a shared vector space and retrieves the $k$ most similar items by cosine distance or approximate nearest neighbours \citep{lewis2020rag,borgeaud2022improving,ram2023incontext}. Dense passage retrieval \citep{karpukhin2020dpr} and contrastive pre-training \citep{izacard2022unsupervised} have substantially improved retrieval quality within this paradigm.

The fundamental limitation is architectural: similarity retrieval assumes that useful memories are representationally close to the query. This holds for many practical cases --- finding relevant documents for a factual question --- but fails precisely where biological memory excels. The smell of smoke does not resemble a campfire visually, yet one retrieves the other. Stairs approaching a landing do not resemble a slip, yet the association is immediate. Similarity-based systems cannot recover these cross-modal, cross-context associations when the memories are distant in embedding space, because the retrieval mechanism requires the very proximity that is absent.

Recent work on hypothetical document embeddings \citep{gao2023precise} and query rewriting \citep{ma2023queryrewriting} attempts to bridge this gap by transforming queries to be closer to target documents, but the retrieval mechanism remains similarity-based --- the transformation must guess the right direction in advance.

Recent systematic surveys confirm that reasoning failures in large language models are particularly acute in embodied and causal domains where grounded experience is absent \citep{song2026reasoning}, reinforcing that text-space similarity alone cannot substitute for associations formed through lived experience.

\subsection{Learned Memory Addressing}

Memory Networks \citep{weston2015memory,sukhbaatar2015e2e}, Neural Turing Machines \citep{graves2014ntm}, and Differentiable Neural Computers \citep{graves2016hybrid} introduce learned read/write operations over external memory. The addressing mechanism --- whether content-based, location-based, or a combination --- is trained end-to-end on downstream task loss.

These architectures demonstrate that learned memory addressing is feasible and powerful. However, the associative structure that emerges is a side effect of task optimisation, not a representation of the temporal structure of experience. A Memory Network trained on question answering learns to address memories that are useful for answering questions. It does not learn that stairs are associated with slipping because those events co-occurred in the agent's experience --- it would only learn this association if it appeared in the training data as a question-answer pair.

The distinction matters for open-ended agents. An embodied system accumulating experience needs memory associations that reflect the structure of that experience, not the structure of any particular downstream task. Our approach learns associative structure from temporal co-occurrence alone, making the resulting memory useful for any downstream purpose without task-specific training.

\subsection{Attractor Dynamics and Hopfield Networks}

Classical Hopfield networks \citep{hopfield1982neural} and their modern continuous-state successors \citep{krotov2016dense,krotov2021large,ramsauer2021hopfield} provide elegant content-addressable memory through energy-based pattern completion. Given a partial or noisy version of a stored pattern, the network converges to the nearest stored attractor.

Modern Hopfield networks — developed from the polynomial interaction functions of \citep{krotov2016dense}, to the exponential formulation of \citep{ramsauer2021hopfield} — have been connected to attention mechanisms in transformers, establishing a formal link between energy-based memory and self-attention. This connection is valuable for understanding intra-sequence memory, but the framework encodes no structure \emph{between} stored patterns. Each memory is an isolated attractor; the energy landscape defines basins of attraction around individual patterns but no associative paths between them.

Our work is complementary. Hopfield-style pattern completion could serve as the mechanism by which a partial state cue activates a full stored memory within a single attractor basin. The Inward predictor then provides the between-memory structure: which other stored patterns are associatively reachable from the completed pattern. The two mechanisms operate at different levels --- pattern completion within, associative traversal between.

\subsection{Graph-Based Relational Memory}

Knowledge graphs \citep{bordes2013transe,sun2019rotate}, graph neural networks over relational data \citep{schlichtkrull2018rgcn}, and recent work on graph-based retrieval \citep{edge2024graphrag} represent relational structure explicitly. Nodes are entities, edges are typed relations, and retrieval is graph traversal.

The power of graph-based approaches --- and the reason they capture something that flat similarity retrieval cannot --- is precisely that they encode \emph{structure between} entities. ``Paris is-capital-of France'' captures a relation that embedding similarity alone might not prioritise. In this sense, graph-based memory shares our motivation: relational structure matters.

The limitation is that this structure typically requires explicit construction pipelines. Knowledge graph construction requires entity extraction, relation typing, and coreference resolution --- a pipeline that introduces its own failure modes \citep{ji2022survey}. Recent work on automated graph construction from corpora \citep{edge2024graphrag} reduces but does not eliminate this dependency: the extraction system still determines which relations are expressible. Temporal co-occurrence --- the training signal we propose --- requires no extraction pipeline. Two states are associated if they co-occurred within the same temporal window. The structure is learned directly from the stream of experience, not extracted from it by a separate system.

\subsection{Predictive Representations and the Successor Representation}

The successor representation (SR) \citep{dayan1993successor} learns a predictive map of expected future state occupancy under a given policy: $M(s, s') = \mathbb{E}[\sum_{t=0}^{\infty} \gamma^t \mathbf{1}[s_t = s'] \mid s_0 = s]$. Deep extensions \citep{kulkarni2016deep,barreto2017successor} scale this to continuous state spaces. The SR is the closest computational predecessor to our Inward predictor in that it learns \emph{predictive} structure from \emph{temporal} experience.

The key distinction is directional. The SR predicts \emph{forward} occupancy --- where the agent will go next, along trajectories. It is a model of the future under a policy. The Inward predictor predicts \emph{lateral} association --- which past states across the agent's entire history are experientially linked to the current state, regardless of trajectory direction or policy. The SR would not associate the approach to a staircase with a slip that happened \emph{earlier} on different stairs; the Inward predictor would, because both were experienced within the same temporal window as structurally similar events. Additionally, the SR is typically tied to a specific policy; our associations are policy-independent, reflecting the structure of experience itself.

\subsection{Spreading Activation, Cognitive Models, and Neuroscience of Episodic Memory}

\citet{collins1975spreading} proposed spreading activation as the mechanism for semantic memory retrieval: activating a concept spreads activation to associated concepts via weighted links in a semantic network. The model accounts for priming effects \citep{meyer1971facilitation}, typicality gradients, and the fan effect. It remains the dominant cognitive model of semantic memory.

Anderson's ACT-R \citep{anderson1993rules,anderson2004integrated} formalised spreading activation within a broader cognitive architecture, incorporating base-level activation (recency and frequency), contextual activation from current goals, and retrieval thresholds. The base-level activation equation captures something close to our familiarity normalisation: memories that are constantly active receive a high baseline, and only deviations above baseline drive retrieval.

From neuroscience, Tulving's \citeyearpar{tulving1972episodic,tulving2002episodic} distinction between episodic and semantic memory established that retrieving a specific past experience is fundamentally different from retrieving general knowledge --- a distinction that maps directly onto our Inward (episodic association) versus Outward (semantic similarity) predictors. \citeauthor{eichenbaum2004hippocampus}'s \citeyearpar{eichenbaum2004hippocampus} relational memory theory argues that the hippocampus specifically encodes the relationships \emph{between} elements of experience, not the elements themselves --- it is an organ for learning associative structure. Hippocampal replay during sleep and rest \citep{wilson1994reactivation,carr2011hippocampal} provides a biological mechanism for the transitive bridging we propose: replayed sequences from separate episodes can overlap, creating associative links between states that were never directly co-experienced. This is precisely the mechanism by which our iterative predictor traversal bridges separate trajectories.

The insight that memory retrieval is \emph{traversal of associative structure} rather than \emph{search by similarity} is the correct one. What spreading activation lacks is a mechanism for \emph{learning} the associative structure from experience. The semantic network is hand-built. Link weights are manually assigned or estimated from behavioural data. The model describes the dynamics of retrieval but not the dynamics of \emph{how the associative structure was acquired}.

Our Inward predictor provides a computational implementation of this missing piece. The predictor is trained on temporal co-occurrence --- the same signal that, in a biological system, would produce the associative links that spreading activation traverses. The predictor's forward pass \emph{is} the spreading activation, implemented as a differentiable neural computation that can be trained end-to-end on the structure of experience. Iterative application of the predictor (using the predicted region as input for a subsequent prediction) implements multi-hop spreading activation, enabling the transitive bridging that accounts for creative recombination.

We note that our approach belongs to the broader family of predictive processing theories \citep{rao1999predictive,clark2013whatever}, and shares the prediction-error minimisation ethos of active inference \citep{friston2010freeenergy}. However, active inference addresses action selection under a generative model --- the \emph{use} of predictions for planning --- whereas PAM addresses the \emph{structure} of associative memory from which predictions are drawn. The two are complementary: an active inference agent could use PAM-style associative retrieval as the memory substrate over which it plans.

\subsection{Positioning}

Table~\ref{tab:positioning} summarises the positioning.

\begin{table*}[t]
\centering
\caption{Positioning of PAM relative to existing memory architectures.}
\label{tab:positioning}
\resizebox{\textwidth}{!}{%
\begin{tabular}{@{}lccccccc@{}}
\toprule
Property & \begin{tabular}[c]{@{}c@{}}Similarity\\Retrieval\end{tabular} & \begin{tabular}[c]{@{}c@{}}Memory\\Networks\end{tabular} & \begin{tabular}[c]{@{}c@{}}Hopfield\\Nets\end{tabular} & \begin{tabular}[c]{@{}c@{}}Knowledge\\Graphs\end{tabular} & \begin{tabular}[c]{@{}c@{}}Successor\\Rep.\end{tabular} & \begin{tabular}[c]{@{}c@{}}Spreading\\Activation\end{tabular} & \textbf{PAM (Ours)} \\
\midrule
Learned structure & $\times$ & Task-supervised & $\times$ & Extracted & Fwd.\ temporal & $\times$ & \textbf{Lateral temporal} \\
Between-memory relations & $\times$ & Implicit & $\times$ & Explicit & Forward only & Explicit (hand-built) & \textbf{Learned, bidirectional} \\
Cross-context association & $\times$ & If in task data & $\times$ & If extracted & $\times$ (policy-bound) & If in graph & \checkmark \\
Transitive bridging & $\times$ & $\times$ & $\times$ & Multi-hop queries & Via occupancy & Spreading & \textbf{Iterative prediction} \\
No task supervision & \checkmark & $\times$ & \checkmark & $\times$ & \checkmark & \checkmark & \checkmark \\
No explicit construction & \checkmark & \checkmark & \checkmark & $\times$ & \checkmark & $\times$ & \checkmark \\
Differentiable traversal & N/A & \checkmark & \checkmark & $\times$ (discrete) & \checkmark & $\times$ (discrete) & \checkmark \\
\bottomrule
\end{tabular}
}%
\end{table*}


\section{Architecture}

\subsection{Background: Joint-Embedding Predictive Architectures}

A Joint-Embedding Predictive Architecture (JEPA) \citep{lecun2022path} learns representations by predicting in \emph{latent space} rather than in observation space. Given two views of the same data (e.g., two temporal crops of a video), an encoder maps each view to a representation, and a predictor network learns to predict one representation from the other. Crucially, the target representation is produced by a slowly-updated copy of the encoder (via exponential moving average), with gradients stopped through the target branch. This stop-gradient/EMA mechanism prevents representational collapse --- the trivial solution where all inputs map to the same point --- by creating a moving target that the predictor must continuously chase. I-JEPA \citep{assran2023ijepa} and V-JEPA \citep{bardes2024vjepa} demonstrated this approach for images and video respectively, learning rich semantic representations without pixel-level reconstruction.

PAM extends this framework by applying the same prediction-in-latent-space mechanism to a new target domain: not future sensory states as a JEPA does but associatively linked past states stored in a memory bank.

\subsection{Overview}

The system consists of two complementary JEPA architectures operating over the same embedding space but with different target domains:

\begin{itemize}[leftmargin=*]
\item \textbf{Outward JEPA}: Given current sensory state, predict the next sensory state. This is the standard JEPA formulation \citep{lecun2022path,assran2023ijepa,bardes2024vjepa} --- a world model that predicts forward in real time. Importantly, the Outward JEPA is not memory-free: it continuously builds and updates a model of relationships between entities in the world (spatial layouts, functional properties, causal regularities). This world model constitutes a form of semantic memory --- knowledge about how things generally behave --- that complements the episodic associations learned by the Inward JEPA.

\item \textbf{Inward JEPA}: Given current composite state, predict which \emph{past} states are associatively reachable. This is the episodic memory system --- a predictor that navigates the associative structure of stored experience.
\end{itemize}

Both share a common encoder that maps multimodal experience into a shared embedding space. They differ in what they predict: the Outward predictor targets future states in the sensory stream; the Inward predictor targets temporally co-occurring past states in the memory bank. The Outward predictor's world model provides the similarity structure (what things are like each other); the Inward predictor provides the associative structure (what things were experienced together). Both contribute to memory, but through different mechanisms.

\subsection{Composite State Representation}

At each timestep $t$, the encoder produces a composite state embedding:

\begin{equation}
s(t) = f_\theta(x_v(t), x_a(t), x_p(t), x_c(t))
\end{equation}

where $x_v$, $x_a$, $x_p$, $x_c$ denote visual, auditory, proprioceptive, and contextual inputs respectively, and $f_\theta$ is the encoder network. The composite state captures not just which objects are present but the full multimodal context --- body position, sounds, the action being performed. Two moments involving the same objects in different activities produce different composite states, eliminating the need for explicit task labels.

\subsection{Temporal Co-occurrence as Training Signal}

The Inward predictor is trained on temporal co-occurrence. Given a state $s(t)$, the training targets are states from the temporal neighbourhood:

\begin{equation}
\mathcal{N}_\tau(t) = \{s(t') : |t - t'| \leq \tau, \; t' \in \mathcal{H}\}
\end{equation}

where $\tau$ is the temporal window and $\mathcal{H}$ is the agent's history. The predictor learns to map $s(t)$ to a region of embedding space that contains the embeddings of states in $\mathcal{N}_\tau(t)$.

Critically, this is not similarity learning. Two states can be temporally co-occurring (and thus training targets for each other) without being representationally similar. The approach to a staircase and the moment of slipping are maximally associated (temporal co-occurrence $= 1$ within the episode) while potentially distant in embedding space.

\paragraph{Scope and extensions.} We emphasise that temporal co-occurrence is not the \emph{only} signal that shapes biological memory --- it is the \emph{foundational} signal that this study isolates. In biological systems, emotional arousal, causal inference, and semantic relatedness all modulate association strength. A single traumatic event creates strong associations without repeated co-occurrence; the emotional signal acts as a multiplier on the encoding threshold. In the full architecture we envision, affective and arousal signals would enter as additional dimensions of the composite state --- effectively, another input channel to the Inward predictor --- amplifying the co-occurrence signal for high-salience experiences. We expect these affective signals to emerge from the architecture's own dynamics (e.g., prediction error magnitude, novelty relative to the world model, goal-state discrepancy) rather than being hand-designed. The present work deliberately isolates the temporal co-occurrence primitive to establish what associative structure it alone can support. Salience-weighted co-occurrence is an extension that strengthens the foundation without changing it.

\subsection{The Inward Predictor}

The Inward predictor $g_\phi$ takes a composite state and produces a predicted target representation:

\begin{equation}
\hat{z}(t) = g_\phi(s(t))
\end{equation}

The training objective minimises the distance between the predicted representation and the actual representations of temporally co-occurring states, while maximising distance from non-co-occurring states:

\begin{equation}
\mathcal{L}_{\text{assoc}} = \sum_{t^+ \in \mathcal{N}_\tau(t)} d(\hat{z}(t), \bar{s}(t^+)) - \lambda \sum_{t^- \notin \mathcal{N}_\tau(t)} d(\hat{z}(t), \bar{s}(t^-))
\end{equation}

where $\bar{s}$ denotes stop-gradient targets (following JEPA convention to prevent representational collapse) and $d$ is a distance function in embedding space.

\paragraph{Geometric interpretation.} The predictor output $\hat{z}(t)$ is a \emph{point} in embedding space --- the predicted centroid of the association neighbourhood. It does not explicitly represent a distribution or region boundary. The ``region'' around this point is defined implicitly by the retrieval threshold $\epsilon$ (Section~3.6). A richer formulation --- predicting both a mean and variance to represent association uncertainty as a Gaussian in embedding space --- is a natural extension that we leave to future work. The point-prediction formulation is sufficient to demonstrate the core claim that learned associative structure captures information inaccessible to similarity retrieval.

The stop-gradient on target states is essential. Without it, the encoder could collapse all temporally co-occurring states to a single point --- technically achieving zero prediction loss while destroying all representational structure. The exponential moving average (EMA) update of the target encoder provides the moving target that forces the predictor to learn genuine associative structure.

\subsection{Memory Retrieval}

Given a query state $s_q$ (typically the current sensory state), retrieval is:

\begin{enumerate}[leftmargin=*]
\item Compute predicted association target: $\hat{z} = g_\phi(s_q)$
\item Retrieve stored states whose embeddings fall within the predicted region: $\mathcal{R} = \{s_i \in \mathcal{M} : d(\hat{z}, \bar{s}_i) < \epsilon\}$
\end{enumerate}

where $\mathcal{M}$ is the memory bank and $\epsilon$ is a retrieval threshold. This is a single forward pass through the predictor followed by a nearest-neighbour lookup in the predicted region --- computationally equivalent to similarity retrieval but navigating learned associative structure rather than representational distance. As with standard similarity retrieval, the nearest-neighbour lookup scales to large memory banks via approximate nearest-neighbour indices \citep{johnson2019billion}; the predictor simply redirects which point the index is queried from.

\subsection{Transitive Bridging and Creative Recombination}

Multi-hop retrieval is implemented as iterated prediction. Given query state $s_q$:

\begin{enumerate}[leftmargin=*]
\item First hop: $\hat{z}_1 = g_\phi(s_q)$ $\rightarrow$ retrieve states $\mathcal{R}_1$
\item Second hop: For each $s_i \in \mathcal{R}_1$, compute $\hat{z}_2^{(i)} = g_\phi(s_i)$ $\rightarrow$ retrieve states $\mathcal{R}_2$
\item Continue for $k$ hops.
\end{enumerate}

If trajectory 1 visits states A$\rightarrow$B and trajectory 2 visits B$\rightarrow$C, then from state A, the first hop reaches B (co-occurred in trajectory 1) and the second hop reaches C (co-occurred in trajectory 2). State C is retrieved despite never having co-occurred with A. This is the mechanism we term \emph{creative bridging} --- novel recombination grounded in actual experience. It requires that states B appear in both trajectories; Section~5.5 tests whether this condition holds in our benchmark.

The combinatorial complexity is bounded but rich. If the memory bank has $N$ states and each has $K$ associations, single-hop retrieval accesses $O(K)$ states, two-hop accesses $O(K^2)$, and $k$-hop accesses $O(K^k)$. This exponential expansion is what gives rise to the richness of associative memory, but every path in the expansion is grounded in actual temporal co-occurrence --- it is recombination, not hallucination.

\subsection{Familiarity Normalisation}

In a static environment where objects co-occur constantly, raw temporal co-occurrence becomes uninformative. Association strength must be normalised against a familiarity baseline:

\begin{equation}
w_{\text{assoc}}(s_i, s_j) = w_{\text{raw}}(s_i, s_j) - \mathbb{E}[w_{\text{raw}}(s_i, s_j)]
\end{equation}

where the expectation is estimated as a running average of co-occurrence frequency. Only the residual above baseline produces a meaningful association.

This is sensory adaptation applied to memory formation. An organism stops noticing constant stimuli; analogously, constantly co-occurring states stop producing strong specific associations. The mechanism is self-calibrating: in a novel environment (a holiday), the baseline is near zero and every co-occurrence is salient --- explaining the subjective density of holiday memories. Upon return to a familiar environment, the baseline re-establishes and encoding density drops.

\subsection{Adaptive Decay}

As the memory bank grows, the association graph becomes denser and the signal-to-noise ratio degrades. Adaptive decay ties the forgetting rate to memory accumulation:

\begin{equation}
\lambda_{\text{decay}}(t) = f(\bar{\rho}(t))
\end{equation}

where $\bar{\rho}(t)$ is the average association density at time $t$. As the graph densifies, decay accelerates to maintain retrieval precision. This provides bounded memory capacity with unbounded experience --- natural forgetting as a homeostatic mechanism rather than an information loss.

\subsection{\texorpdfstring{Similarity $\times$ Association = Specificity}{Similarity x Association = Specificity}}

The dual architecture produces a principled account of episodic specificity. The Outward JEPA learns \emph{similarity}: states that share functional properties cluster together in embedding space. The Inward JEPA learns \emph{association}: states that co-occurred temporally are linked regardless of their representational distance.

Three retrieval modes emerge from the interaction:

\begin{itemize}[leftmargin=*]
\item \textbf{Similarity without association}: ``That looks like a drill.'' Recognition without episodic context. The Outward encoder activates and the Inward predictor does not reach a specific memory.
\item \textbf{Association without similarity}: ``Something about this moment reminds me of Tuesday.'' Priming without identified content. The Inward predictor reaches a region and the Outward encoder cannot resolve what.
\item \textbf{Both together}: ``That's the drill that stripped the screw on Tuesday.'' Full episodic recall with object grounding. Both channels converge on the same target, dividing the search space and providing specificity.
\end{itemize}

The specificity emerges from intersection, not union. Where similarity retrieval alone returns all drills, and association retrieval alone returns everything from Tuesday, their intersection returns \emph{that specific drill on that specific occasion}. This is an architectural prediction: episodic specificity requires both channels, and systems with only similarity retrieval (RAG) or only associative retrieval will produce qualitatively different --- and less specific --- retrieval patterns.


\section{Experimental Setup}

We evaluate PAM on a synthetic benchmark designed to test the \emph{faithfulness of associative recall} --- how accurately the predictor recalls associations that were actually experienced --- rather than generalisation to unseen associations. This distinction is fundamental to the evaluation paradigm and requires explanation.

\subsection{Recall vs Retrieval: The Evaluation Paradigm}

\paragraph{Evaluation contract.} PAM is evaluated as an episodic recall system, not a retrieval generalisation system. The question is: given associations the agent actually experienced, does it faithfully reactivate them? This differs from standard retrieval benchmarks, where held-out generalisation is a desired property. In our setting, strong performance on unexperienced associations is not required --- indeed, it would indicate false-link formation. Accordingly, we train on the full experienced association set and evaluate recall fidelity, specificity, and false-association rate.

Standard retrieval systems (RAG, dense passage retrieval) are evaluated on generalisation: given a concept of relevance, find relevant items the system was not explicitly trained on. This is the correct paradigm for document retrieval, where the system should generalise from seen queries to unseen ones.

Associative memory is \emph{recall rather than retrieval}. An association is a specific link formed by specific experience: the agent saw the hammer with the nail on Tuesday, so they are associated. Testing whether the system retrieves associations it never formed is testing whether someone remembers events they never witnessed. The correct evaluation asks: given that the system experienced these temporal co-occurrences, does it faithfully recall them?

The failure modes for associative recall are therefore different from those for retrieval:

\begin{itemize}[leftmargin=*]
\item \textbf{False association} (hallucinating links that never existed) rather than failure to generalise
\item \textbf{Loss of specificity} (retrieving whole categories instead of specific experiences) rather than low recall
\item \textbf{Signal corruption} (confusing temporal structure with embedding geometry) rather than distribution shift
\end{itemize}

Our metrics are designed to measure these failure modes directly.

\subsection{Synthetic World}

The world consists of 20 rooms in a 128-dimensional embedding space. Each room is defined by a cluster centroid with scale 2.0; rooms are well-separated so that states within the same room are representationally similar while states in different rooms are representationally distant. Fifty objects are distributed across rooms (5 per room, with some objects shared between rooms). Objects contribute features with scale 1.5, modulating the room embedding.

Experience is generated as 500 trajectories of 100 timesteps each (50,000 total states). At each timestep, the agent occupies a room and perceives the objects present, producing a composite state embedding. Trajectories traverse multiple rooms, creating temporal co-occurrences between states from different rooms --- the critical property that allows us to test whether the predictor can recall associations that cross representational boundaries.

Temporal associations are defined by a co-occurrence window of $\tau = 5$ timesteps, yielding 242,264 total associations. Because trajectories cross room boundaries, some of these associations link states from different rooms that share no embedding similarity.

\subsection{Models}

\paragraph{Predictor (PAM).} A 4-layer MLP with residual connections: 128 $\rightarrow$ 1024 $\rightarrow$ 1024 $\rightarrow$ 1024 $\rightarrow$ 128, with GELU activations, residual connections in layers 2--3, and layer normalisation on the output (2.36M parameters). The architecture is a feedforward network rather than a transformer because the Inward predictor operates on \emph{single composite states}, not sequences --- there is no sequence to attend over. A transformer would be appropriate if the predictor processed trajectories as input; the current formulation maps a single state to a single predicted point, for which an MLP is the natural architecture. Trained with InfoNCE loss using in-batch negatives (i.e., all other items in the batch serve as negatives for each positive pair; batch size 512, yielding 511 negatives per positive), cosine learning rate schedule ($5 \times 10^{-4}$ $\rightarrow$ $1 \times 10^{-5}$), and temperature annealing ($0.15 \rightarrow 0.05$) over 500 epochs. InfoNCE implements the contrastive objective described in Section~3.5 in a probabilistic form: it maximises the log-probability of the true associate relative to in-batch negatives, which is equivalent to minimising distance to positives while maximising distance to negatives. Because the embedding space is fixed in these experiments (no encoder to collapse), the stop-gradient and EMA mechanisms described in Section~3.5 are not required here; they become necessary when the Inward predictor is integrated with a co-evolving encoder (Section~6.8). The predictor is trained on \emph{all} temporal associations in the experience stream, consistent with the recall paradigm: we are training the system on its complete experience and testing whether it faithfully recalls what it experienced.

Note that we evaluate the predictor component in isolation, over a fixed synthetic embedding space, rather than as part of a full JEPA system with joint encoder training. In a full JEPA, the encoder learns representations from raw sensory data while the predictor learns to predict in that evolving latent space. Here, the synthetic world provides the embedding space directly --- there is no raw data for an encoder to process. This isolation is deliberate: it allows us to attribute results to the Inward predictor's learned associative structure without confounding from encoder-level representation effects. The predictor is architecturally identical to a JEPA predictor; the difference is that its input space is fixed rather than co-evolving. Integration with a learned encoder (e.g., V-JEPA over video) is the natural next step (Section~6.8).

\paragraph{Cosine similarity baseline.} Standard $k$NN retrieval by cosine distance in the embedding space. No training required --- this represents the similarity-only approach used by RAG and dense retrieval systems.

\paragraph{Bilinear similarity baseline.} A learned bilinear form $s(x, y) = x^T W y$ trained on the same temporal co-occurrence data as the predictor, using the same InfoNCE objective. The bilinear baseline is included as a form-of-solution test --- whether a single learned linear compatibility function suffices to recover cross-boundary temporal associations --- not as a parameter-matched competitor. It tests whether the \emph{form} of the mapping matters (linear vs nonlinear), not the capacity. The bilinear model converges during training and achieves above-chance overall discrimination (AUC = 0.791), indicating that it learns within-room associative structure where the linear transformation can exploit geometric proximity between associates. However, on cross-room pairs it drops to chance (AUC = 0.514) and scores zero on cross-boundary recall and specificity (Table~\ref{tab:results}). The nonlinear predictor is necessary specifically to learn associations that cross representational boundaries.

\subsection{Metrics}

We define five metrics designed for the recall paradigm. The architecture (Section~3.6) describes retrieval as finding states within an $\epsilon$-ball around the predicted point. For evaluation, we use equivalent top-$k$ ranking by distance from the predicted point, which avoids threshold sensitivity and enables standardised comparison across methods. The two formulations rank identically; top-$k$ simply fixes the retrieval window size rather than the distance threshold.

\paragraph{Cross-Boundary Recall@$k$.} For query states paired with associated targets from \emph{different rooms} (where cosine similarity provides zero signal), what fraction of true associations appear in the top-$k$ predictions? This is the core thesis metric: can the predictor recall associations that cross representational boundaries?

\paragraph{Association Precision@$k$.} Of the top-$k$ retrieved states, what fraction are \emph{actual} temporal associates of the query (from any room)? This measures false association rate --- the recall analogue of hallucination.

\paragraph{Discrimination AUC.} Area under the ROC curve for the binary classification: ``were these two states experienced within the same temporal window?'' This measures the predictor's overall ability to separate experienced-together from never-experienced-together states, regardless of retrieval rank.

\paragraph{Specificity@$k$.} Among retrieved states, what fraction are the \emph{correct specific} associates rather than merely same-category members? This measures episodic specificity --- the difference between ``a drill'' and ``that drill from Tuesday.''

\paragraph{Cross-Room Discrimination AUC.} Discrimination AUC restricted to cross-room pairs, isolating the predictor's ability to discriminate in the regime where embedding similarity is uninformative.

\paragraph{Formal definitions and query selection.} All metrics are macro-averaged (per-query scores computed independently, then averaged across queries) to ensure robustness across queries with varying association richness. For Cross-Boundary Recall@$k$, let $\mathcal{A}^{\times}(q)$ denote a query's cross-room temporal associates. We compute $\text{CBR@}k(q) = |\text{top-}k \cap \mathcal{A}^{\times}(q)| / |\mathcal{A}^{\times}(q)|$, macro-averaged over 500 queries with $|\mathcal{A}^{\times}(q)| \geq 3$. For Association Precision@$k$, $\text{AP@}k(q) = |\text{top-}k \cap \mathcal{A}(q)| / k$ where $\mathcal{A}(q)$ includes all temporal associates; macro-averaged over 500 queries with $|\mathcal{A}(q)| \geq 3$. Discrimination AUC uses the Wilcoxon--Mann--Whitney U-statistic per query: $\text{AUC}(q) = \frac{1}{n_+ n_-} \sum_{i \in \text{pos}} \sum_{j \in \text{neg}} \mathbb{1}[s_i > s_j] + \tfrac{1}{2}\mathbb{1}[s_i = s_j]$, with negatives subsampled to 2,000 when $n_- > 2{,}000$; macro-averaged over 300 queries with $|\mathcal{A}(q)| \geq 5$. Cross-Room Discrimination AUC restricts positives to $\mathcal{A}^{\times}(q)$, computed over the subset of AUC queries with $|\mathcal{A}^{\times}(q)| \geq 3$ (188 queries). For Specificity@$k$, let $\mathcal{D}(q)$ be the category distractors (same-room states that are not true associates of $q$); $\text{Spec@}k(q) = h_{\text{true}} / (h_{\text{true}} + h_{\text{dist}})$ where $h_{\text{true}}$ and $h_{\text{dist}}$ are cross-room associates and category distractors in the top-$k$ respectively; macro-averaged over up to 300 queries. All query selections use a fixed random seed for reproducibility.

\subsection{Ablation Controls}

Two controls validate that the learned signal reflects genuine temporal structure:

\paragraph{Temporal shuffle control.} We randomly permute the temporal ordering within each trajectory while preserving all state embeddings. This destroys temporal co-occurrence structure while maintaining the embedding geometry. If the predictor has learned temporal structure, shuffling should collapse performance; if it has learned embedding geometry artifacts, shuffling should have no effect.

\paragraph{Similarity-matched negatives.} We construct negative pairs from same-room, same-category states that were never temporally co-present. These distractors are maximally similar in embedding space but lack the temporal co-occurrence link. If the predictor discriminates true associates from these matched negatives, it has learned episodic specificity --- not merely category membership.

\subsection{Training Details}

Training was performed on a single NVIDIA RTX 4080 Super (CUDA 12.4). Training time is approximately 350 seconds. The model was trained on all 242,264 temporal associations, consistent with the recall paradigm. The synthetic world, training, and evaluation code are available at \url{https://github.com/EridosAI/PAM-Benchmark}. All experiments use deterministic random seeds (world generation: seed 42; multi-seed evaluation: seeds 42, 123, 456) for full reproducibility.


\section{Results}

\subsection{Faithful Recall Across Representational Boundaries}

Table~\ref{tab:results} presents the primary results.

\begin{table*}[t]
\centering
\caption{Faithfulness of associative recall (mean $\pm$ SD across training seeds 42, 123, 456). The predictor is the only method that recalls associations across embedding space boundaries. Both baselines achieve above-chance overall discrimination by exploiting within-room geometric proximity, but drop to chance on cross-room pairs and score zero on cross-boundary recall --- confirming that cross-boundary association requires nonlinear transformation of the temporal co-occurrence signal.}
\label{tab:results}
\small
\begin{tabular}{@{}lccc@{}}
\toprule
Metric & Predictor & Cosine Similarity & Bilinear \\
\midrule
Association Precision@1 & $\mathbf{0.970 \pm 0.005}$ & 0.000 & 0.015 \\
Association Precision@5 & $\mathbf{0.703 \pm 0.001}$ & 0.085 & 0.037 \\
Association Precision@20 & $\mathbf{0.216 \pm 0.001}$ & 0.045 & 0.022 \\
Cross-Boundary Recall@20 & $\mathbf{0.421 \pm 0.002}$ & 0.000 & 0.000 \\
Discrimination AUC (overall) & $\mathbf{0.916 \pm 0.000}$ & 0.789 & 0.791 \\
Discrimination AUC (cross-room) & $\mathbf{0.849 \pm 0.004}$ & 0.503 & 0.514 \\
Specificity@20 & $\mathbf{0.338 \pm 0.005}$ & 0.000 & 0.000 \\
\bottomrule
\end{tabular}
\end{table*}

\begin{figure}[t]
\centering
\includegraphics[width=\columnwidth]{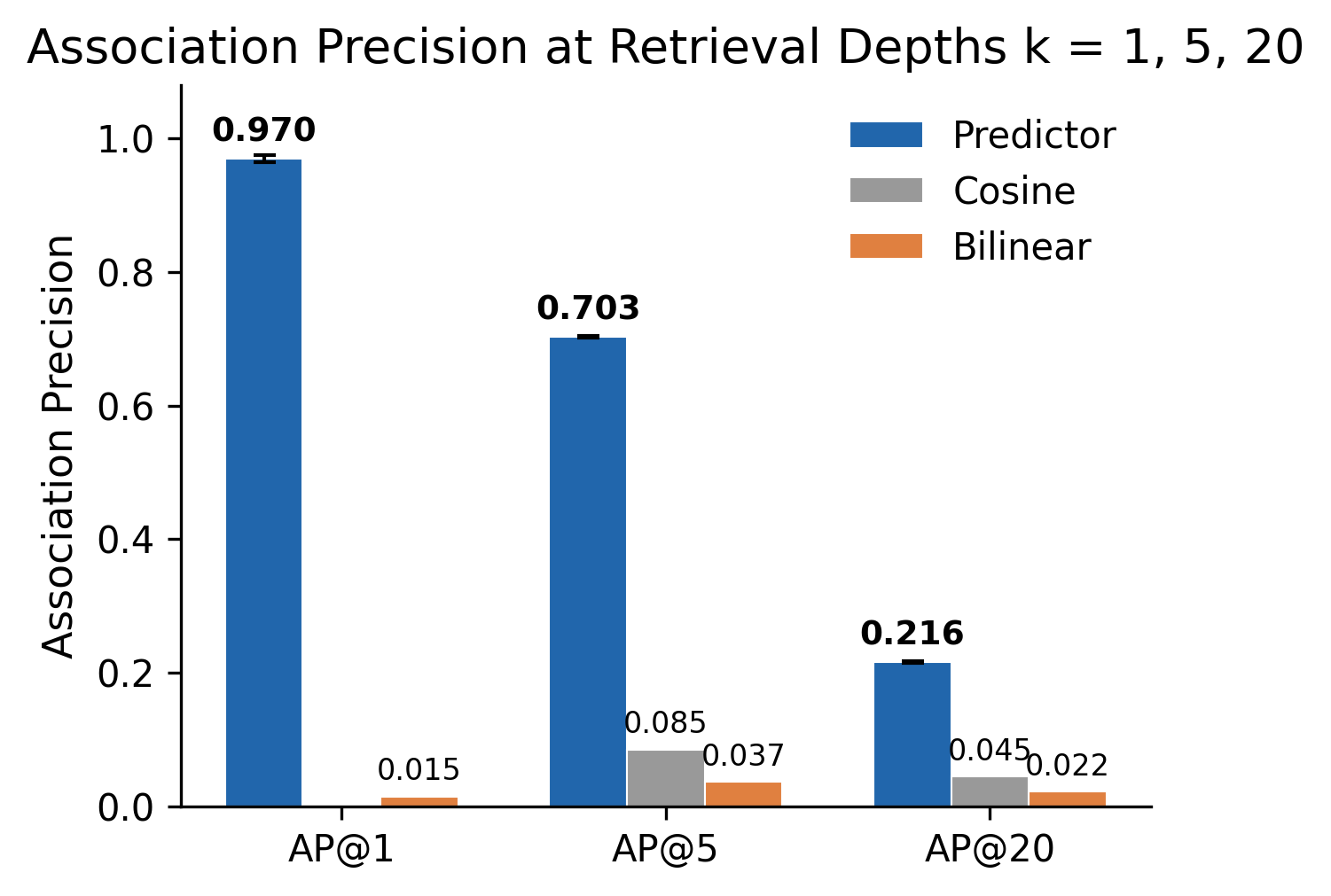}
\caption{Association Precision at retrieval depths $k = 1, 5, 20$. The predictor's top retrieval is a true associate 97\% of the time; precision dilutes at greater depth as the retrieval window extends beyond the tight association neighbourhood. Cosine and bilinear baselines score near zero at all depths. Error bars show $\pm 1$ SD across three training seeds.}
\label{fig:precision}
\end{figure}

\begin{figure}[t]
\centering
\includegraphics[width=\columnwidth]{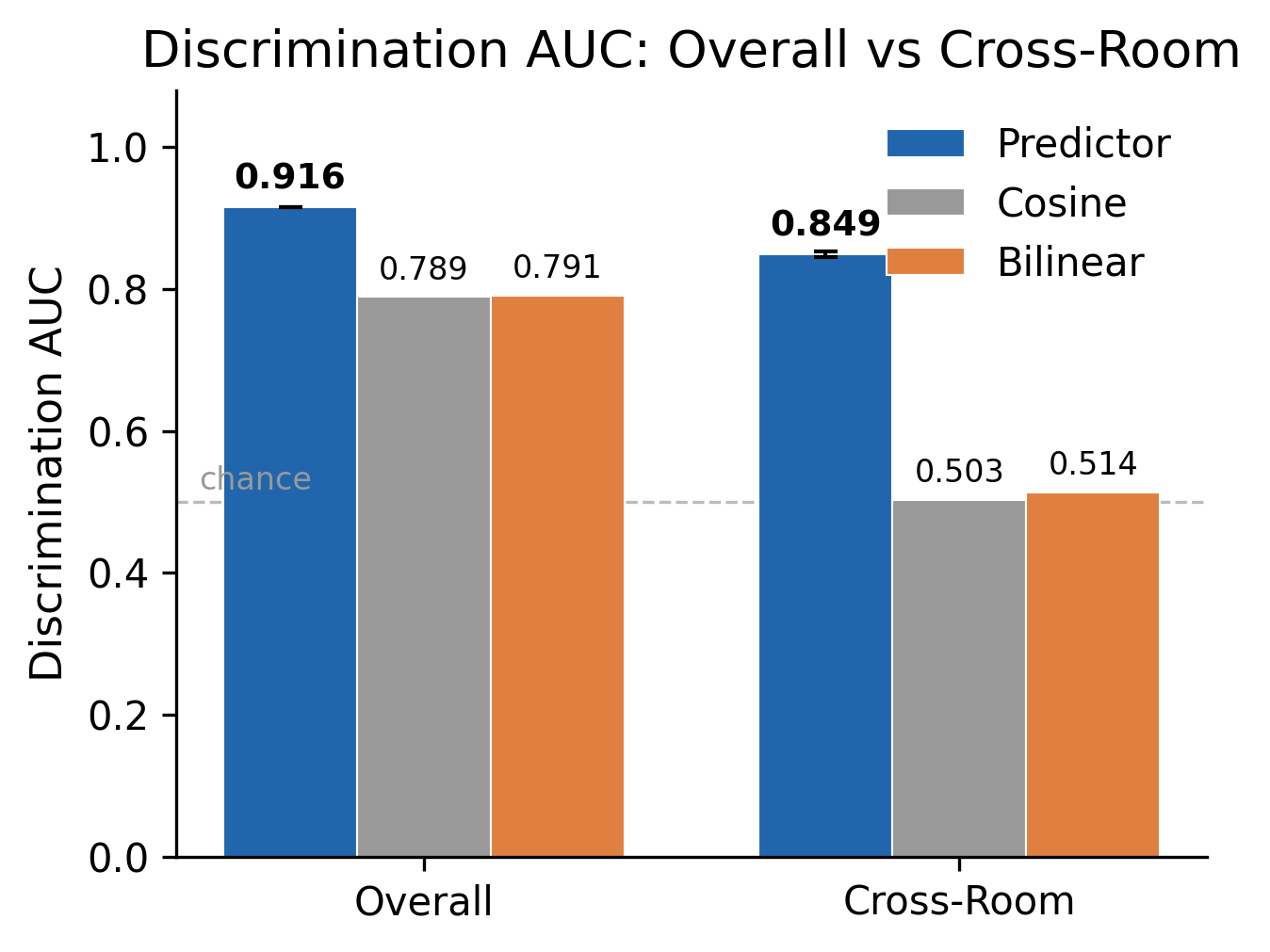}
\caption{Discrimination AUC --- overall vs cross-room. All three methods achieve above-chance overall discrimination by exploiting within-room geometric proximity. On cross-room pairs --- where the thesis claim lives --- cosine and bilinear drop to chance (dashed line) while the predictor maintains AUC = 0.849.}
\label{fig:discrimination}
\end{figure}

The headline result is the precision gradient across retrieval depths. The predictor's top retrieval is a genuine temporal associate 97\% of the time (AP@1 = 0.970). At depth 5, precision remains strong at 0.703 --- on average, approximately 3.5 of the top 5 retrievals are true associates. By depth 20, precision falls to 0.216. This gradient reveals that the predictor learns a well-ordered ranking --- true associates are concentrated at the top of the retrieval list, with precision diluting as the retrieval window expands into the tail of the predicted region. The decline from AP@1 to AP@20 reflects a calibration problem in tail retrieval, not a failure of core association learning: the predictor identifies the right target but the predicted region is broader than the true association neighbourhood, admitting non-associated states at greater depth.

Cross-Boundary Recall@20 = 0.421 with cosine at zero demonstrates that the predictor recalls associations across representational boundaries --- the regime where similarity-based methods fundamentally cannot operate. This gap is architectural rather than quantitative, in the well-separated embedding geometry tested here: the information cosine similarity requires (representational proximity) does not exist for cross-room pairs. In continuous embedding spaces without clean cluster boundaries, the gap may narrow (Section~6.7). Note that cosine's precision increases with depth (AP@1 = 0.000, AP@5 = 0.085, AP@20 = 0.045): its nearest neighbour is the most geometrically similar state, which is rarely a temporal associate, but at greater retrieval depth some incidental within-room associates are captured.

The discrimination AUC of 0.916 means the predictor separates ``experienced together'' from ``never experienced together'' with 91.6\% accuracy across all pairs. Cosine similarity achieves 0.789 on overall discrimination --- reflecting its ability to identify within-room associates via geometric proximity --- but drops to chance (0.503) on cross-room pairs where embedding similarity is uninformative. The predictor maintains AUC = 0.849 even on cross-room pairs, demonstrating that it has learned genuine associative structure beyond what embedding geometry provides.

All metrics vary by less than $\pm 0.006$ (SD) across three training seeds (42, 123, 456), suggesting that results are robust to training seed variation rather than dependent on favourable initialisation. Results are additionally stable under alternative query selections: resampling the query set with five different random seeds produces SD $\leq 0.012$ on all primary metrics (AP@1 SD = 0.012, CBR@20 SD = 0.010, AUC SD = 0.007, AP@20 SD = 0.003), with the default seed 42 values falling within one standard deviation of the resample means in all cases.

\subsection{Ablation Controls}

\paragraph{Temporal shuffle control.} Randomly permuting temporal ordering within trajectories (while preserving all embeddings) collapses Cross-Boundary Recall@20 from 0.421 to 0.044 --- a 90\% reduction. This confirms that the predictor has learned genuine temporal co-occurrence structure, not artifacts of the embedding geometry. The residual 0.044 likely reflects incidental geometric regularities that correlate with the shuffled pseudo-associations. The effect replicates across training seeds: on seed 123, temporal shuffling produces an equivalent collapse (CBR@20 $-$90\%, AP@20 $-$92\%), confirming that the ablation result is robust and not seed-dependent.

\begin{figure}[t]
\centering
\includegraphics[width=\columnwidth]{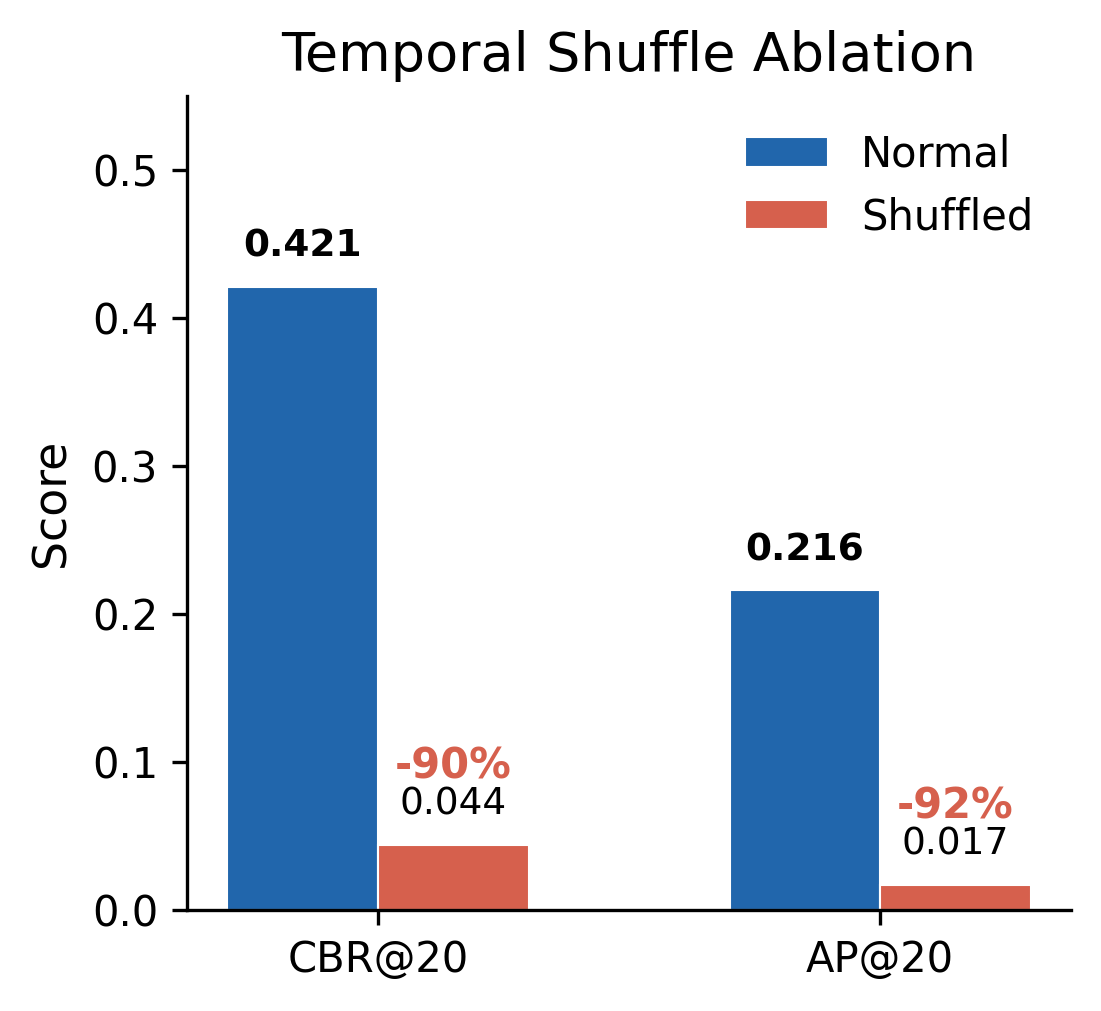}
\caption{Temporal shuffle ablation. Randomly permuting temporal order within trajectories collapses cross-boundary recall by 90\% and association precision by 92\%, confirming that the predictor learned genuine temporal co-occurrence structure.}
\label{fig:shuffle}
\end{figure}

\paragraph{Similarity-matched negatives.} When negative pairs are drawn from same-room, same-category states that were never temporally co-present, the predictor achieves AUC = 0.848 versus cosine's 0.732. The predictor correctly discriminates actual temporal associates from similar-but-not-associated distractors. This proves episodic specificity: the predictor has learned something beyond ``these states are from the same category.'' It has learned \emph{which specific states within a category were experienced together}.

The cosine baseline's AUC of 0.732 (above chance) on this metric reflects the fact that within-room associates do share some geometric similarity --- temporal neighbours within the same room are slightly closer in embedding space than arbitrary same-room states, because they share temporal context features. The predictor's 0.848 substantially exceeds this, confirming that temporal co-occurrence provides discriminative signal beyond what embedding similarity captures even within the same room.

\subsection{Held-Out Query-State Evaluation}

A natural concern is that the predictor, trained on all associations, has simply learned a lookup table mapping each query embedding to its target neighbourhood, an embedding shortcut rather than a structural understanding of associative relationships. To test this, we held out 20\% of query states from training (no associations anchored at these states appeared in training data) and evaluated recall using these held-out states as queries. Held-out states were excluded only from the anchor (input) side of training pairs; they could still appear as targets in other states' association pairs. The predictor has therefore seen these embeddings as outputs but has never learned to route \emph{from} those specific inputs. Under the recall evaluation contract (Section~4.1), collapse on unseen anchors is an expected signature of anchor-specific episodic recall rather than a failure mode.

\begin{table}[t]
\centering
\small
\caption{Held-out query-state evaluation.}
\label{tab:heldout}
\begin{tabular}{@{}lc@{}}
\toprule
Condition & CBR@20 \\
\midrule
Train-anchor queries & 0.508 \\
Held-out queries & 0.000 \\
\bottomrule
\end{tabular}
\end{table}

The collapse is total: held-out queries produce zero cross-boundary recall. The predictor has learned a query-specific mapping, it recalls associations from perspectives it actually experienced, not from novel viewpoints.

\begin{figure}[t]
\centering
\includegraphics[width=\columnwidth]{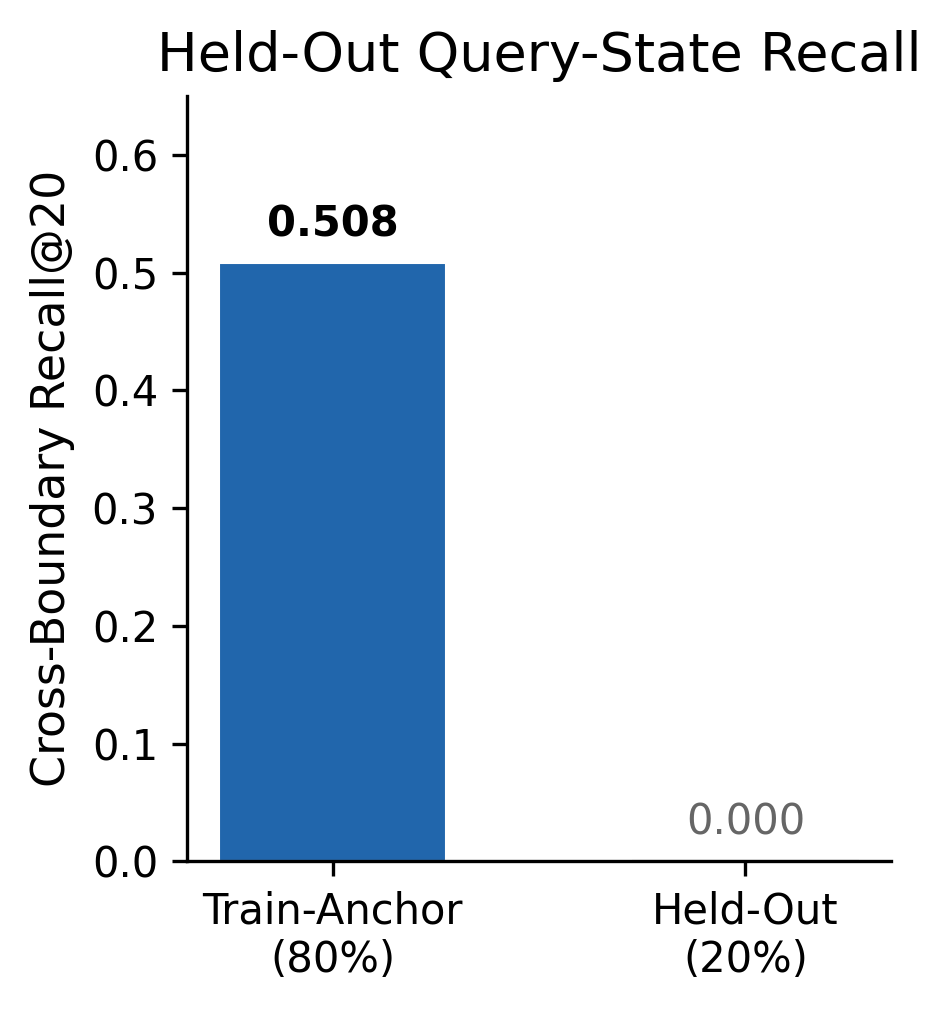}
\caption{Held-out query-state evaluation. Train-anchor queries achieve CBR@20 = 0.508; queries never used as training anchors score zero. The predictor recalls associations from experienced viewpoints only, consistent with anchor-specific episodic recall.}
\label{fig:heldout}
\end{figure}

This result is consistent with the episodic character of the memory system. Episodic memory is inherently perspective-bound: you recall what Tuesday's workshop looked like from where you stood, not from an angle you never occupied. The predictor has learned specific $f_\theta(\text{query}) \rightarrow \text{target}$ mappings for each experienced anchor state. This \emph{is} faithful memorisation of experienced associations --- the system remembers what it experienced, from the perspectives at which it experienced it. A system that produced strong recall for viewpoints it never anchored would be hallucinating associations, not remembering them.

The elevated train-anchor CBR@20 (0.508 vs 0.421 in the full evaluation) reflects the reduced query pool: the held-out split removes 20\% of queries, and the remaining queries may have slightly richer association neighbourhoods on average.

\subsection{Generalisation Stress Test}

As a secondary analysis, we evaluated the predictor under a standard retrieval paradigm: 70/30 edge-disjoint train/test split on association edges.

\begin{table}[t]
\centering
\small
\caption{Generalisation stress test: recall of held-out associations.}
\label{tab:generalisation}
\begin{tabular}{@{}lcc@{}}
\toprule
Condition & R@20 & Cosine R@20 \\
\midrule
Train associations & 0.578 & 0.000 \\
Held-out associations & 0.023 & 0.000 \\
\bottomrule
\end{tabular}
\end{table}

The large train-test gap (0.578 vs 0.023) is expected and correct under the recall paradigm. The system was trained to recall experienced associations; associations it never experienced are, by definition, outside its memory. Testing held-out associations is testing whether someone remembers conversations they never had.

The non-zero held-out performance (0.023 vs cosine's 0.000) is a secondary finding: some structural regularity in the association graph transfers beyond memorised pairs. The predictor has learned not only specific associations but some of the underlying temporal structure that generates them --- consistent with the emergent transitive internalisation observed in Section~5.7, where the predictor reaches multi-hop targets in a single pass. This suggests partial structural generalisation, though the primary claim remains faithful recall rather than generalisation.

\subsection{Creative Bridging: Boundary Identification}

We designed a creative bridging test to evaluate cross-trajectory recombination: recalling state C via the chain A$\rightarrow$B$\rightarrow$C where A--B and B--C occurred in separate trajectories linked by a shared object at B.

Oracle analysis revealed that the task is structurally unsolvable in our synthetic world. The temporal association graph contains zero cross-trajectory edges --- associations are defined strictly within trajectories. An oracle with perfect knowledge of all bridge states reaches the target 0\% of the time.

This negative result identifies a boundary condition for the temporal co-occurrence primitive: creative recombination requires \emph{entity persistence across episodes}. In our synthetic world, trajectories sample independently. In real embodied experience, the same physical drill appears on Tuesday and Thursday, creating overlapping temporal neighbourhoods that would produce cross-trajectory edges. The creative bridging mechanism described in Section~3.7 is theoretically grounded but requires entity persistence that our benchmark does not provide. Empirical validation requires either a redesigned synthetic world or a real experience stream, both of which we pursue in ongoing work with the dual-JEPA architecture.

\subsection{Model Selection Diagnostics}

The final configuration (D2) was selected through systematic ablation during development. Table~\ref{tab:diagnostics} presents the key experiments using cross-room R@20 and MRR --- retrieval-paradigm metrics used for architecture search prior to the recall-paradigm evaluation reported above. These metrics guided model selection but are not the primary evaluation; they are reported here for reproducibility and to document the design rationale.

\begin{table*}[t]
\centering
\caption{Architecture selection diagnostics (development-phase metrics). R@20 for configs Baseline through D is T1\_R@20 evaluated on a 70/30 train split; the D2 value (0.421) is CBR@20 from the full faithfulness evaluation (Section~5.1), reported here because D2 was selected as the final configuration and evaluated under the paper's primary metrics. The values track consistently across evaluation protocols.}
\label{tab:diagnostics}
\small
\begin{tabular}{@{}llcccccc@{}}
\toprule
Config & & Layers & Hidden & Pairs & Sampling & R@20 & MRR \\
\midrule
Baseline & & 3 & 256 & 100k & Fixed & 0.037 & 0.060 \\
+ Capacity & & 3 & 1024 & 100k & Fixed & 0.218 & 0.530 \\
+ Depth (B) & & 4 & 1024 & 100k & Fixed & 0.222 & 0.567 \\
+ Coverage (C) & & 3 & 1024 & 200k & Fixed & 0.305 & 0.469 \\
+ Both, online (D) & & 4 & 1024 & 200k & Online & 0.398 & 0.423 \\
\textbf{+ Both, fixed (D2)} & & \textbf{4} & \textbf{1024} & \textbf{200k} & \textbf{Fixed} & \textbf{0.421} & \textbf{0.635} \\
\bottomrule
\end{tabular}
\end{table*}

\begin{figure}[t]
\centering
\includegraphics[width=\columnwidth]{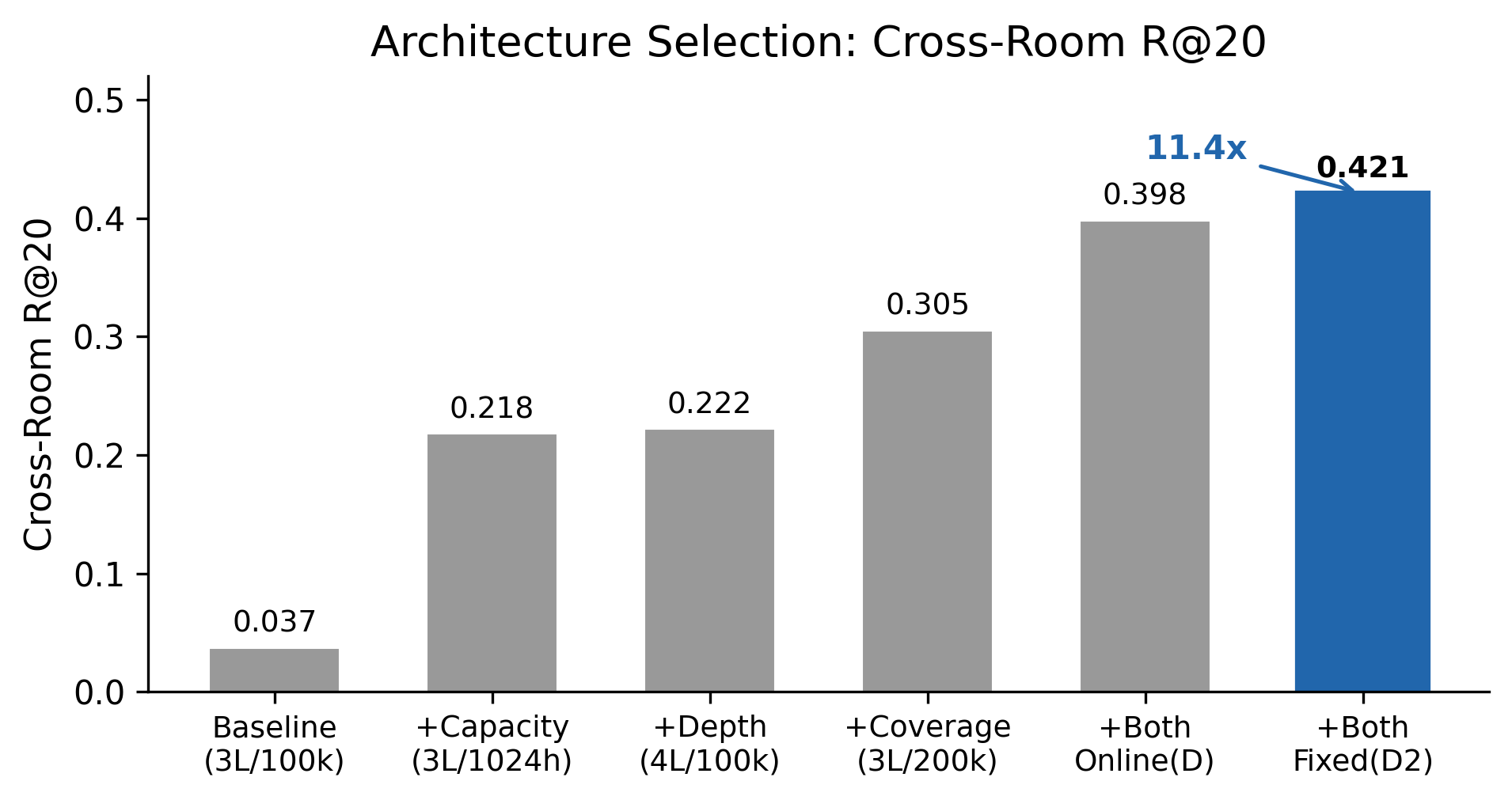}
\caption{Architecture selection progression. Cross-room R@20 improves 11.4$\times$ from baseline to final configuration (D2), driven primarily by data coverage and the interaction of capacity with coverage.}
\label{fig:model_selection}
\end{figure}

Three findings emerge:

\paragraph{Data coverage was the primary bottleneck.} The world contains 242,264 temporal associations. At 100k training pairs, only 41\% are covered --- the system literally cannot recall associations it was never exposed to. Increasing to 200k pairs (82\% coverage) nearly doubled R@20 from 0.218 to 0.305 with the same architecture. Under the recall paradigm, this is expected: broader experience produces richer memory.

\paragraph{Capacity and coverage interact multiplicatively.} With limited data, added capacity provides minimal benefit (+2\%). With broader coverage, the deeper network extracts substantially more value (+30\%). Neither factor alone accounts for the full gain.

\paragraph{Fixed pairs outperform online sampling.} Online sampling (fresh pairs each epoch) converges poorly (loss 2.315 vs 0.409) and MRR drops from 0.635 to 0.423. This aligns with the recall paradigm: temporal associations are \emph{specific facts to be consolidated}, not a distribution to be approximated. The fixed-pair regime mirrors biological memory consolidation through repetition --- repeated exposure to the same association strengthens recall, exactly as the complementary learning systems framework predicts \citep{mcclelland1995cls}.

\subsection{Supplementary Findings}

Two additional analyses provide supporting evidence for properties discussed in the architecture section. These were conducted during development using cross-room R@20 as the evaluation metric and a single training seed; the qualitative patterns are consistent with the main results but have not been replicated under the full multi-seed evaluation protocol.

\paragraph{Transitive chain traversal.} The predictor sustains cross-room recall across multi-hop association chains: 1-hop R@20 = 0.455, 2-hop = 0.355, 3-hop = 0.280 (cosine: 0.000 at all depths). The monotonic degradation confirms learned transitive structure. Notably, 1-hop cross-room recall (0.455) exceeds the overall CBR@20 (0.421), suggesting the predictor has partially internalised transitive associations --- reaching some 2-hop targets in a single forward pass without explicit iterative traversal. This mirrors the cognitive phenomenon of mediated priming, where strong associative chains produce direct priming without conscious intermediate steps \citep{balota1986depth,mcnamara1992theories}.

\begin{figure}[t]
\centering
\includegraphics[width=\columnwidth]{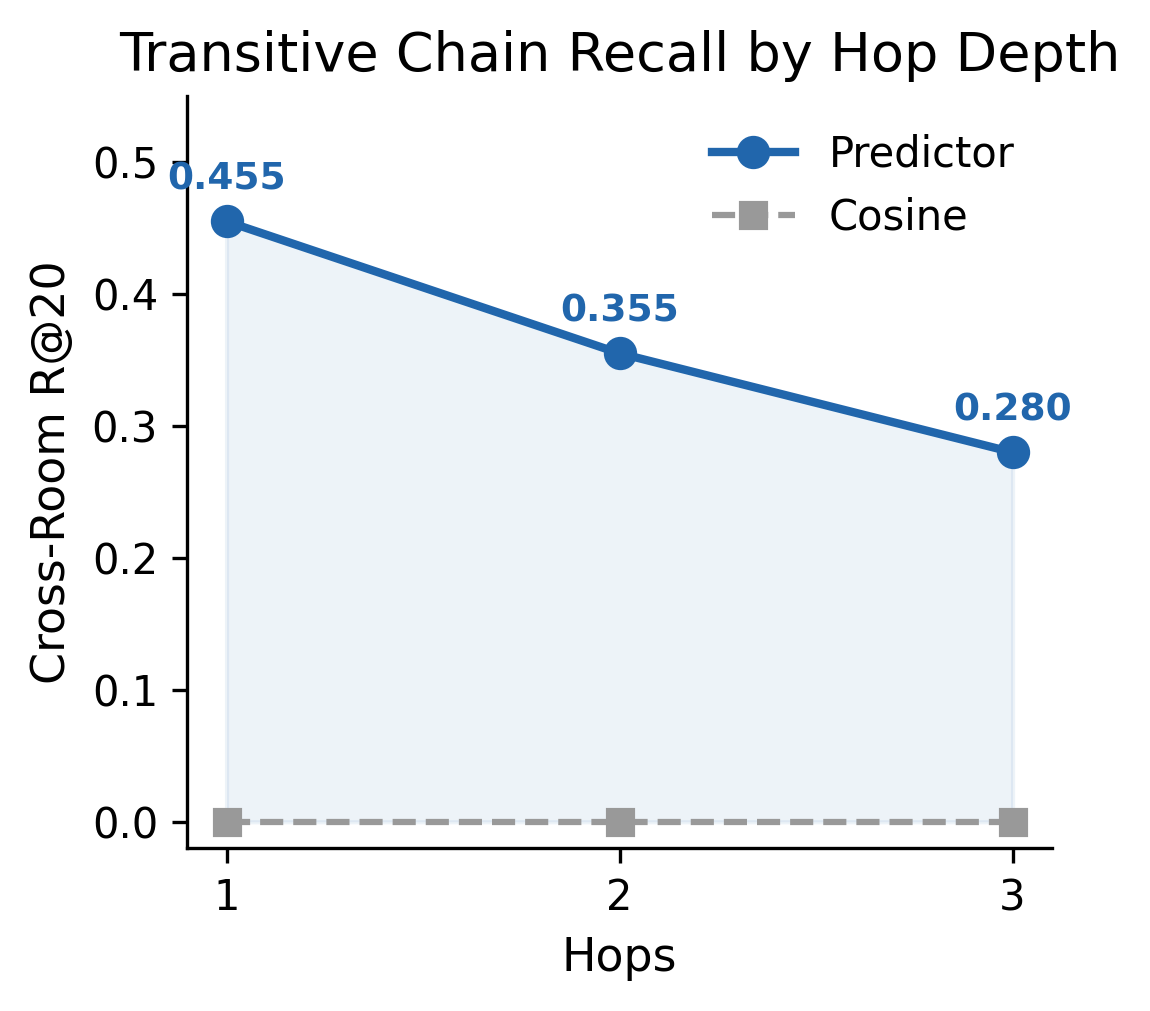}
\caption{Transitive chain recall by hop depth. The predictor sustains cross-room recall across multi-hop association chains with graceful degradation, while cosine similarity scores zero at all depths.}
\label{fig:transitive}
\end{figure}

\paragraph{Adaptive decay scaling.} Recency-weighted retrieval improves mean precision by +43.2\% relative to uniform retrieval on the current best configuration (with-decay: 0.607, without: 0.424). The improvement confirms that adaptive decay provides meaningful signal-to-noise benefit even at the modest scale of the synthetic world (50k states, Section~3.9). The relatively contained margin reflects the strength of the 4-layer predictor: the model's improved capacity raises the no-decay baseline substantially. Theoretical considerations predict that the decay advantage should grow with memory bank size as the association graph densifies --- a prediction consistent with earlier experiments on larger worlds, but not formally validated at scale with the current architecture.


\section{Discussion}

\subsection{Implications for RAG and Similarity-Based Systems}

The structural zero for cosine similarity on cross-boundary recall has direct implications for Retrieval-Augmented Generation and similar systems. RAG retrieves by embedding similarity, which means it systematically misses an entire class of useful associations: those linked by temporal co-occurrence rather than representational proximity. A RAG system asked ``What might go wrong on wet stairs?'' will retrieve documents about stairs, about wetness, and about accidents --- but not the specific memory of nearly slipping on marble stairs in a hotel lobby ten years ago. That memory shares no representational features with the current query; the link exists only in the temporal structure of experience.

The discrimination AUC results (0.916 overall, 0.849 cross-room vs cosine's 0.789 overall, 0.503 cross-room) quantify this gap. On cross-room pairs --- where the thesis claim lives --- the predictor separates true associates from non-associates with 85\% accuracy while cosine similarity is at chance. This is not a quantitative improvement amenable to better embeddings or more data --- it is a qualitative capability that pure similarity-based retrieval does not acquire without incorporating a temporal training signal. Hybrid approaches that augment similarity with query rewriting or temporal context (e.g., HyDE; \citealp{gao2023precise}) may partially bridge this gap, but they inject temporal information extrinsically rather than learning associative structure from experience. For embodied agents accumulating continuous experience, this capability enables grounded recall across modalities and episodes --- a prerequisite for long-horizon planning and personalised memory that similarity-based retrieval alone cannot provide.

\subsection{Memorisation as Correct Behaviour}

The finding that fixed training pairs outperform online sampling, and that the train/test gap under edge-disjoint splitting is large (0.578 vs 0.023), might appear concerning under a retrieval evaluation paradigm. Under the recall paradigm, both results are expected and correct.

Temporal associations are specific facts formed by specific experience. The goal of an associative memory system is to recall these facts faithfully --- not to generalise to associations the system never formed. A biological memory system that ``generalised'' to events it never witnessed would be \emph{hallucinating}, not remembering. The correct evaluation of associative memory asks: given that you experienced this, do you recall it accurately?

This changes how associative memory systems should be trained. The fixed-pair regime succeeds because repeated exposure to the same association strengthens recall --- mirroring biological memory consolidation through rehearsal and replay \citep{wilson1994reactivation}. Online sampling treats associations as a distribution to be approximated, which is the wrong inductive bias for episodic memory. This asymmetry --- episodic systems benefit from repetition, semantic systems from variety --- is a prediction of the dual-JEPA framework and an instance of the complementary learning systems division.

\subsection{Episodic Specificity: Demonstrated at the Within-Room Level}

The similarity-matched negatives control (Section~5.2) provides the first empirical evidence for the episodic specificity claim. The predictor discriminates true temporal associates from same-room, same-category distractors (AUC = 0.848 vs cosine's 0.732). This means the predictor has learned not just ``these states are from the same category'' but ``these specific states were experienced together.''

This is a component of the full specificity claim made in Section~3.10 --- that the intersection of Outward (similarity) and Inward (association) produces episodic specificity. The present result demonstrates the Inward component in isolation: given states that are already similar (same room), the predictor still discriminates based on temporal co-occurrence. The full dual-channel intersection --- where both similarity and association converge on the same target --- remains an architectural prediction to be tested with the integrated Outward JEPA.

\subsection{The Temporal Shuffle as Validation}

The 90\% collapse in cross-boundary recall under temporal shuffling (0.421 $\rightarrow$ 0.044) is the most important control in the paper. It definitively rules out the possibility that the predictor has learned an artifact of the embedding geometry rather than genuine temporal structure.

The obvious alternative explanation is that the predictor has simply learned to map states to some geometric property that happens to correlate with room transitions, rather than learning the actual temporal co-occurrence structure. The temporal shuffle control tests exactly this hypothesis. By preserving all state embeddings while destroying temporal structure, it isolates the predictor's dependence on temporal signal. The 90\% collapse confirms: the predictor learned temporal co-occurrence, not embedding geometry.

\subsection{Creative Bridging: An Honest Negative}

The creative bridging null result (Section~5.5) identifies a precise boundary condition: creative recombination through temporal co-occurrence requires entity persistence across episodes. Without persistent entities, the association graph contains no cross-trajectory edges, and no traversal mechanism --- however sophisticated --- can bridge experiences that share no common states.

This clarifies what the temporal co-occurrence primitive alone can and cannot do. Within a continuous experience stream, it supports faithful recall and transitive bridging. Across disconnected episodes, it requires the additional structure that entity persistence provides. In real embodied experience (a workshop with persistent tools), this structure exists naturally. In our synthetic benchmark, it does not.

We additionally identified an architectural constraint: the predictor maps each query to a single point in embedding space, but cross-trajectory associations are inherently one-to-many. Addressing this may require mixture-of-experts outputs or hybrid retrieval. Creative recombination remains a theoretically grounded prediction of the architecture; its empirical validation is the subject of ongoing work with entity-persistent environments and the full dual-JEPA system.

\subsection{The Outward JEPA's Role in Memory}

This paper has focused on the Inward predictor in isolation, but the full architecture includes an Outward JEPA that contributes its own form of memory. The Outward predictor continuously builds a world model --- learning which states tend to follow which, what objects afford, how spatial layouts constrain movement. This constitutes semantic memory: general knowledge about how things behave, independent of any specific episode.

The interaction between similarity (Outward) and association (Inward) is where the architecture predicts full episodic specificity will arise (Section~3.10). We have demonstrated the Inward component's contribution to specificity (Section~6.3). Evaluating the full interaction --- where both predictors operate over a shared, co-evolving embedding space --- is the natural next step. The architecture additionally supports imagination (reaching regions beyond stored states through learned transition patterns) and creative recombination (transitive bridging across episodes with persistent entities), though these capabilities remain theoretical predictions of the framework rather than demonstrated results.

\subsection{Limitations}

\paragraph{Synthetic benchmark.} The world is small (20 rooms, 50 objects, 50k states) and geometrically clean. Real experience streams are high-dimensional, noisy, and lack clear cluster boundaries. The structural gap between predictor and similarity baselines follows from the architecture, but absolute recall rates will differ.

\paragraph{No joint embedding learning.} The embedding space is fixed. In the full architecture, the encoder and both predictors would co-evolve, potentially producing embeddings better-suited for associative recall. The present results are a lower bound.

\paragraph{Single-point prediction.} Adequate for one-to-one associations but inadequate for one-to-many, as the creative bridging test revealed.

\paragraph{Temporal co-occurrence only.} Biological memory is shaped by emotion, arousal, and causal inference in addition to temporal co-occurrence (Section~3.4). The present work isolates the temporal primitive.

\paragraph{Statistical scope.} Results are reported across three training seeds. While the very tight SDs (e.g., AP@5 SD = 0.001) and consistent query resample stability (SD $\leq 0.012$ across five query seeds) suggest robust findings, a larger seed sweep would strengthen confidence, particularly for metrics with greater variance.

\paragraph{Baseline scope.} The current baselines test whether similarity-based retrieval (cosine) and learned linear similarity (bilinear) suffice for associative recall. Future work should compare against temporal co-occurrence graph methods (personalised PageRank, random walks over temporal edges), successor representations \citep{kulkarni2016deep,barreto2017successor} which learn forward predictive structure from temporal experience, temporal graph neural networks trained on the same co-occurrence edges, and capacity-matched nonlinear baselines to isolate the contribution of PAM's specific architecture from general nonlinear learning.

\paragraph{Noise robustness untested.} The synthetic embedding space is geometrically clean, with well-separated room clusters. Real embedding spaces are continuous and noisy. While the shuffle ablation confirms that the predictor learns temporal structure rather than exploiting geometry, robustness to embedding noise (e.g., Gaussian perturbations of state representations) has not been tested. The structural advantage over similarity baselines follows from the architecture, but absolute performance under noisy embeddings remains an open question.

\paragraph{Association precision at depth.} While the predictor's top retrieval is a true associate 97\% of the time (AP@1), precision at depth 20 is 0.216 --- roughly four in five of the top-20 retrievals are non-associates drawn from the tail of the predicted region. The core prediction is accurate, but the retrieval window at $k = 20$ extends well beyond the tight association neighbourhood. Architectural refinements that produce sharper predicted regions --- variance-aware predictions, adaptive $k$ selection, or confidence-weighted retrieval --- could maintain high precision at greater depth.

\subsection{Future Work}

The most immediate extensions are: (1)~a synthetic world with entity persistence to enable empirical validation of creative bridging; (2)~integration with the Outward JEPA for full dual-architecture evaluation of episodic specificity; (3)~scaling to real multimodal experience streams, initially through an embodied AI platform under development by the authors, where a V-JEPA encoder processes continuous video from a workshop environment.

Longer-term, the emergence of affective signals from the architecture's own dynamics --- prediction error as surprise, goal-state discrepancy as frustration, novelty-familiarity tension as curiosity --- would provide the salience weighting that the present work deliberately omits. These signals, entering as additional input channels to the Inward predictor, would enable one-shot encoding of high-salience events and connect the temporal co-occurrence primitive to the full richness of biological memory formation.


\section{Conclusion}

We have presented Predictive Associative Memory, an architecture in which a JEPA-style predictor learns associative structure from temporal co-occurrence and uses it for faithful recall of experienced associations. We empirically demonstrate associative recall capabilities and provide testable architectural predictions for imagination and cross-episode recombination.

\paragraph{What we showed.}
In this controlled synthetic benchmark, temporal co-occurrence alone is sufficient to learn useful associative recall. The predictor returns a true temporal associate at rank 1 for 97\% of queries ($\mathrm{AP@1}=0.970$). Precision remains strong at depth 5 ($0.703$) and then thins by depth 20 ($0.216$), which is expected as retrieval depth increases.

The key result is cross-boundary behavior. The model recovers associations across representational boundaries where similarity-based retrieval fails outright ($\mathrm{Cross\text{-}Boundary\ Recall@20}=0.421$ vs.\ cosine $=0.000$). It also separates experienced-together from never-experienced-together states with 91.6\% accuracy ($\mathrm{AUC}=0.849$), including cross-room pairs where cosine is near chance. Against similar-but-not-associated distractors, it maintains episodic specificity ($\mathrm{AUC}=0.848$ vs.\ cosine $=0.732$).

Controls indicate the signal comes from temporal structure rather than embedding geometry: temporal shuffling breaks it, and a learned bilinear baseline stays near chance on cross-room discrimination ($\mathrm{AUC}=0.514$) with zero cross-boundary recall. Results are stable across training seeds ($\mathrm{SD}<0.006$) and query selections ($\mathrm{SD}\leq 0.012$). Held-out query-state evaluation shows anchor-specific recall from experienced viewpoints, consistent with episodic memory.

\paragraph{What remains open.}
We have not yet validated full Outward/Inward channel interaction, creative transitive bridging via persistent entities, or imagination-like extrapolation beyond stored states. These are architectural hypotheses for the next phase.

\paragraph{Future work.}
Next steps are to introduce entity-persistent environments, train the dual-channel system jointly, and evaluate on real multimodal streams. Temporal co-occurrence is not a complete memory model, but in this regime it is a sufficient primitive for associative recall that similarity-only methods do not provide.

Temporal co-occurrence is a sufficient foundation for associative recall in the regime tested here. This single primitive, trained without task supervision and without explicit graph construction, enables faithful recall across representational boundaries that similarity-based systems cannot cross. It does not capture everything --- but it captures what similarity alone cannot, and it provides the foundation on which richer memory systems can be built.


\bibliographystyle{abbrvnat}
\bibliography{pam_references}

\end{document}